\documentclass[11pt]{article}

\usepackage[final]{acl}

\usepackage{times}
\usepackage{dsfont}
\usepackage{latexsym}
\usepackage{microtype}
\usepackage{amssymb}
\usepackage{tcolorbox}
\usepackage{amsmath}
\usepackage{fdsymbol}
\usepackage{bbm}
\usepackage{inconsolata}
\usepackage{float}
\usepackage{multirow}
\usepackage{graphicx}
\usepackage{booktabs}
\usepackage{caption}
\usepackage{subcaption}
\usepackage{xspace}
\usepackage{makecell}
\usepackage{enumerate}
\usepackage{booktabs}
\usepackage[capitalize,noabbrev]{cleveref}
\usepackage{tcolorbox}
\usepackage[T1]{fontenc}
\usepackage{listings}
\usepackage{xcolor}

\usepackage[utf8]{inputenc}

\newcommand\eg{\emph{e.g.},\xspace}

\newcommand\aka{\emph{a.k.a.}\xspace}

\newcommand{\papertitle}{\textsc{DeepPlanner}\xspace}

\newcommand{\textttbf}[1]{\texttt{\textbf{#1}}\xspace}
\definecolor{stepcolor}{HTML}{d79b00}
\definecolor{contentcolor}{HTML}{6c8ebf}

\definecolor{cYellow}{RGB}{255,255,3}
\definecolor{cBlue}{RGB}{69,123,157}
\definecolor{cRed}{RGB}{231,56,71}
\definecolor{cRed_1}{RGB}{191,30,46}
\definecolor{cGray}{RGB}{168,218,219}
\definecolor{cBlue_2}{RGB}{5,48,97}
\definecolor{cBlue_1}{RGB}{115,186,214}
\definecolor{cBlue_3}{RGB}{13,76,109}
\definecolor{cBlue_4}{RGB}{64,121,160}
\definecolor{cOrange}{RGB}{250,134,0}
\definecolor{cBlue_6}{RGB}{13,76,109}
\definecolor{cBlue_7}{RGB}{16,106,130}
\definecolor{cBlue_8}{RGB}{19,136,160}
\definecolor{cBlue_9}{RGB}{115,184,214}
\title{\papertitle: Scaling Planning Capability for Deep Research Agents via Advantage Shaping}

\author{
Wei Fan\textsuperscript{1}\thanks{Work done during internship at Amazon.},
Wenlin Yao\textsuperscript{2},
Zheng Li\textsuperscript{2},
Feng Yao\textsuperscript{3},
Xin Liu\textsuperscript{2},
Liang Qiu\textsuperscript{2},\\
\textbf{
Qingyu Yin\textsuperscript{2},
Yangqiu Song\textsuperscript{1},
Bing Yin\textsuperscript{2}}\\
\textsuperscript{1}The Hong Kong University of Science and Technology
\textsuperscript{2}Amazon\\
\textsuperscript{3}University of California, San Diego\\
\texttt{wfanag@connect.ust.hk}, \texttt{fengyao@ucsd.edu}, \texttt{yqsong@cse.ust.hk}\\
\texttt{\{ywenlin,amzzhe,xliucr,liangqxx,qingyy,alexbyin\}@amazon.com}\\
}

\begin{document}
\maketitle
\begin{abstract}
Large language models~(LLMs) augmented with multi-step reasoning and action generation abilities have shown promise in leveraging external tools to tackle complex tasks that require long‑horizon planning.
However, existing approaches either rely on implicit planning in the reasoning stage or introduce explicit planners without systematically addressing how to optimize the planning stage.
As evidence, we observe that under vanilla reinforcement learning~(RL), planning tokens exhibit significantly higher entropy than other action tokens, revealing uncertain decision points that remain under‑optimized.
To address this, we propose \textbf{\papertitle}\footnote{The code and data are available at \url{https://github.com/AlexFanw/DeepPlanner}}, an end-to-end RL framework that effectively enhances the planning capabilities of deep research agents.
Our approach shapes token‑level advantage with an entropy-based term to allocate larger updates to high entropy tokens, and selectively upweights sample-level advantages for planning-intensive rollouts.
Extensive experiments across seven deep research benchmarks demonstrate that \papertitle improves planning quality and achieves state‑of‑the‑art results under a substantially lower training budget.
\end{abstract}

\section{Introduction}
\begin{figure}[t!]
    \centering
    \includegraphics[width=0.47\textwidth]{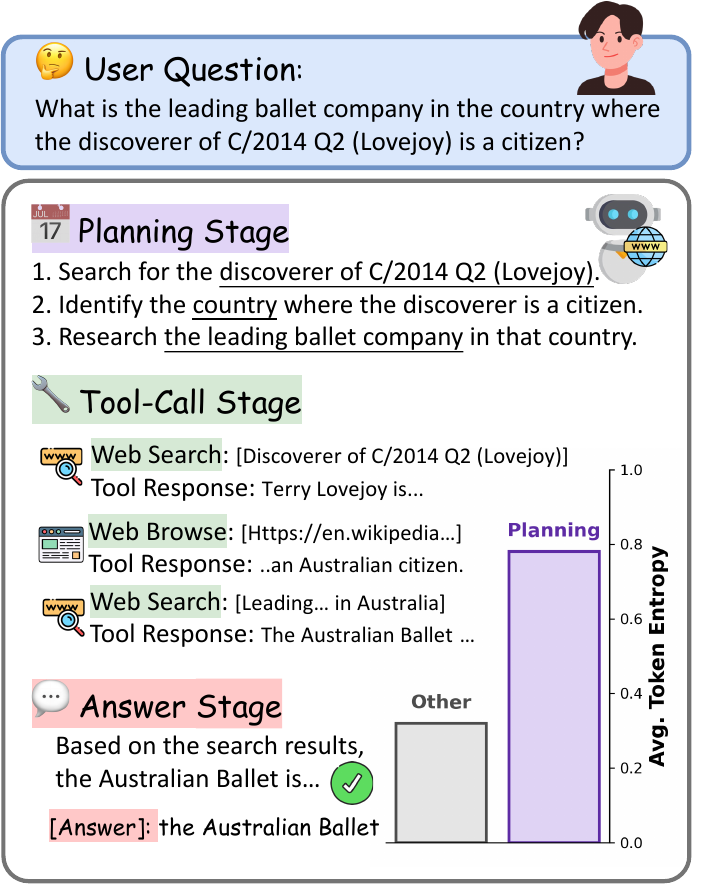}
    \caption{When the agent is instructed to explicitly write a research plan and then execute it, we observe that the planning stage exhibits much higher token entropy (0.78) than other stages (0.32) during RL training.}
    \vspace{-0.2in}
    \label{figs:introduction}
\end{figure}

The evolution of AI assistants has progressed from simple question-answering systems to sophisticated research agents capable of complex information synthesis. While early systems relied solely on the internal knowledge of large language models (LLMs), the introduction of retrieval-augmented generation (RAG)~\cite{lewis2021retrievalaugmentedgenerationknowledgeintensivenlp} greatly expanded their knowledge by incorporating external documents. Deep research ~agents\cite{openaideepresearch,geminideepresearch,grokdeepresearch,zheng2025deepresearcherscalingdeepresearch} represent the next leap in this evolution, combining advanced reasoning capabilities~\cite{deepseekai2025deepseekr1incentivizingreasoningcapability,jaech2024openai} with action generation~\cite{yao2023reactsynergizingreasoningacting} to orchestrate diverse external tools~(\eg web search). These agents automate the complete research workflow of discovering, verifying, and summarizing online information~\cite{xu2025comprehensivesurveydeepresearch}, enabling them to tackle research tasks that traditionally require human expertise.


However, simple reactive approaches that alternate between thinking and acting suffer from inefficiency and goal drift in long-horizon research tasks, necessitating explicit high-level planning to decompose complex goals and coordinate multistage workflows~\cite{huang2024understandingplanningllmagents,wei2025plangenllmsmodernsurveyllm,xu2025comprehensivesurveydeepresearch}.
While some systems like DeepResearcher~\cite{zheng2025deepresearcherscalingdeepresearch} operate without upfront planning, allowing plans to emerge implicitly during reasoning, others, such as \texttt{OpenAI/Agents SDK}\cite{openaisdk} and Cognitive Kernel-Pro\cite{fang2025cognitivekernelproframeworkdeep}, incorporate dedicated planner components for goal decomposition and execution tracking. Despite these framework advances, it still lacks systematic approaches to diagnose and optimize the planning capabilities that are crucial for deep research agents' success.

To diagnose this gap, we extend DeepResearcher~\cite{zheng2025deepresearcherscalingdeepresearch} into a plan-then-execute framework that explicitly decouples high-level planning from low-level execution (\eg tool calls and answer generation).
The agent must propose an initial plan in the first round within \textbf{\texttt{<plan>} \ldots \texttt{</plan>}} and can refine it as new evidence arrives.
Under vanilla Group Relative Policy Optimization (GRPO)~\cite{shao2024deepseekmathpushinglimitsmathematical},
we uncover a consistent pattern shown in ~\cref{figs:introduction,figs:entropy_trends}: planning tokens exhibit substantially higher entropy than other execution tokens throughout training.
Coupled with the performance–entropy transformation mechanism~\cite{cui2025entropymechanismreinforcementlearning}, this observation reveals the challenge that existing methods cannot effectively convert elevated planning-stage entropy into improved downstream performance.

To address this, we introduce \papertitle, an end-to-end reinforcement learning (RL) framework with advantage shaping that sustainably scales the planning capability of deep research agents.
First, inspired by~\cite{cheng2025reasoningexplorationentropyperspective}, we append an entropy-shaped term to the original token-level advantages, amplifying gradients on uncertain tokens (primarily during planning) while clipping to prevent sign flips on strongly negative advantages.
This detached shaping term primarily reinforces advantageous planning trajectories and prevents entropy collapse, preserving sustained exploration.
Second, to further strengthen performance on planning-intensive tasks, we introduce selective advantage upweighting.
Prior work~\cite{zhang2025evolvesearchiterativeselfevolvingsearch} filters rollouts with more tool calls within each RL iteration, then performs supervised fine-tuning (SFT) on these trajectories before continuing RL training.
Instead, within each rollout group under the same query, we identify the most efficient rollout (correct answer, fewest tool calls) and define its tool-call count as the query complexity. We then upweight sample-level advantages for the most efficient rollout in groups exceeding the complexity threshold, achieving comparable gains while maintaining end-to-end simplicity.

Extensive experiments demonstrate that \papertitle achieves state-of-the-art (SOTA) results on deep research benchmarks with markedly fewer training resources: 3,072 queries and 8 rollouts per query, versus 10× more training samples and roughly 2× more rollouts of 
the previous SOTA framework EvolveSearch~\cite{zhang2025evolvesearchiterativeselfevolvingsearch}. Ablations further show that (i) explicit planning improves performance on long-horizon tasks, (ii) entropy-based advantage shaping accelerates effective planning optimization without entropy collapse, and (iii) selective advantage upweighting better exploits complex rollouts that require intensive planning.

Our contributions are summarized as follows:
\begin{itemize}
\item We diagnose and quantify persistent high entropy on planning tokens within a plan-then-execute deep research framework, revealing untapped potential to scale planning capacity.
\item We propose \papertitle, which introduces two advantage shaping mechanisms under GRPO to amplify learning on uncertain planning tokens and on complex, high-quality rollouts, enabling efficient end-to-end training without interleaved SFT.
\item We conduct extensive experiments demonstrating that \papertitle achieves SOTA performance with markedly reduced budgets. Ablations further characterize planning–entropy dynamics and quantify how advantage shaping scales planning capability.
\end{itemize}
\begin{figure*}[t]
    \centering
    \includegraphics[width=2\columnwidth]{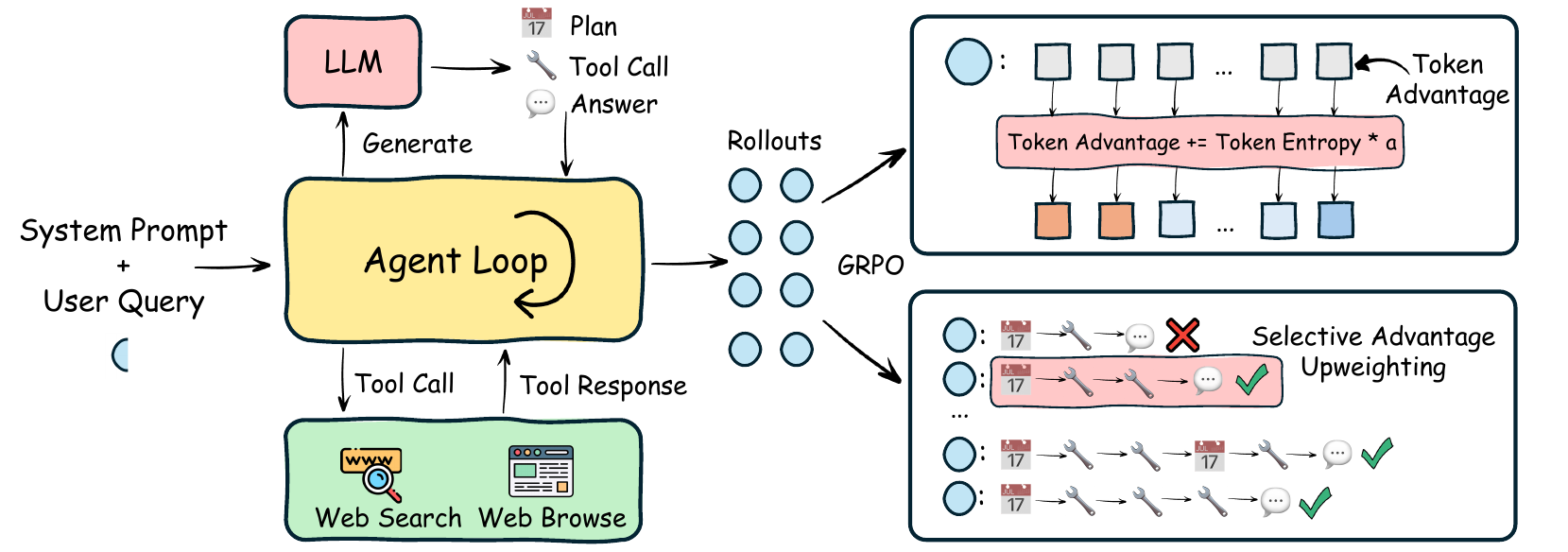}
    \caption{The overview of \papertitle. For each rollout, GRPO token-level advantages are augmented with $\alpha \times$ token entropy (clipped to avoid sign flips). Within each group, rollout(s) that reach the correct answer with fewer tool calls receive a higher weight (tool-call counts gate complexity).}
    \label{figs:method}
    \vspace{-0.2in}
\end{figure*}

\section{\papertitle: Preliminaries}
In this section, we detail the definition of deep research, define the trajectory structure, and describe the basic modules and tools that constitute the environment in which our agent operates.

\subsection{Problem Definition}

In the deep research scenario, given a user query $q$ and system prompt $p$, the LLM-based agent follows the ReAct~\cite{yao2023reactsynergizingreasoningacting} framework to perform multi-turn reasoning (\aka thinking) and action, including \textit{plan}, \textit{tool call}, and \textit{answer}, until producing a final output. The overall process can be represented as a trajectory:
\begin{equation} \label{eq:trajectory}
\small
\tau = \{(s_0, e_0, a_0), (s_1, e_1, a_1), \dots, (s_{T}, e_T, a_{T}), R\},
\end{equation}
where $s_t$ denotes the state at step $t$, consisting of the system prompt $p$, user query $q$, accumulated reasoning~(\aka thinking) tokens $\{e_i\}_{i=0}^{t}$, action tokens $\{a_i\}_{i=0}^{t}$, and tool responses up to $t$. At each step $t$, the model generates (i) a thinking segment $e_t$, and (ii) an action token $a_t$, which correspond to a high-level plan or a low-level execution. $R$ denotes the terminal reward that evaluates the quality of the final answer with respect to the user query.

\subsection{Agent Loop Modules}
At each step, the output of our deep research agent can be decomposed into the following modules:

\paragraph{Think.}  
Before taking any explicit action, we instruct the agent to think and generate a reasoning segment, wrapped in \texttt{<think>} \ldots \texttt{</think>}, serving as the model’s latent observation and analysis about the current state $s_t$.

\paragraph{Plan.}  
The plan is the only high-level action in the trajectory. Unlike prior approaches where planning is often entangled with reasoning or execution, we require the model to explicitly output its intended strategy within \textcolor{purple}{\textbf{\texttt{<plan>} \ldots \texttt{</plan>}}} tags. The plan specifies what information to search, what tools to invoke, and how to integrate the tool response for answering the query.
Two constraints are imposed: (1) an initial plan must be provided in the first round, and (2) the plan can be refined or revisited in later steps. This explicit separation yields interpretable, verifiable plans and discourages opportunistic, ad hoc strategies that can destabilize long-horizon reasoning.

\paragraph{Tool Call.}  
As a low-level action, a tool call can invoke either \texttt{web\_search} or \texttt{web\_browse} external tool. The agent produces a structured request, following a pre-specified tool schema as detailed in~\cref{app:tool_schema}, and encloses it in \texttt{<tool\_call>} \ldots \texttt{</tool\_call>}. The environment then executes the request, returning structured tool outputs as the tool responses, which are appended to the agent context for the next step. 

\paragraph{Answer.}  
The answer module denotes the low-level termination action. When the agent determines that sufficient information has been gathered, it outputs the final response in \texttt{<answer>} \ldots \texttt{</answer>}. Executing this action ends the trajectory, and the agent receives the terminal reward $R$, which reflects answer quality.

\subsection{Deep Research Tools}
During the deep research pipeline, our environment provides two external tools~\cite{zheng2025deepresearcherscalingdeepresearch}:

\paragraph{Web Search.}  
The web search tool is invoked when the agent requires external evidence sources. The agent issues a JSON-formatted request specifying the tool name \texttt{web\_search} and query string(s). The tool calls the online web search API~(Serper\footnote{\url{https://serper.dev/}}) and returns a ranked list of top-$k$ results (we set $k=10$), each containing a \textit{title, URL, and snippet} as follows:
\begin{center}
\vspace{-10pt}
\resizebox{1\linewidth}{!}{
\begin{tabular}{l}
\noindent\textttbf{\textcolor{stepcolor}{Request: }\textcolor{contentcolor}{[query$_1$, query$_2$, ...]}}\\
\noindent\textttbf{\textcolor{stepcolor}{Response: }\textcolor{contentcolor}{(title$_1$, URL$_1$, snippet$_1$), (...}}\\
\end{tabular}
}
\end{center}
Furthermore, both the search query and returned results are cached for downstream use by the web browsing tool.

\paragraph{Web Browse.}  
When a deeper inspection of candidate URLs is needed, the agent calls the browsing tool. The request consists of a target URL together with the associated user query $q$ and search query retrieved from the cache.
The browsing agent first fetches the initial page segment of the document, summarizes query-relevant content, and stores it in short-term memory. Based on this memory, it adaptively decides whether to continue reading subsequent segments of the URL or to stop. Once browsing concludes, the accumulated short-term memory is compiled and returned as the browsing tool response as follows:
\begin{center}
\vspace{-10pt}
\resizebox{1\linewidth}{!}{
\begin{tabular}{l}
\noindent\textttbf{\textcolor{stepcolor}{Request: }\textcolor{contentcolor}{(query, [URL$_1$, URL$_2$, URL$_3$...])}}\\
\noindent\textttbf{\textcolor{stepcolor}{Response: }\textcolor{contentcolor}{(URL$_1$, page information$_1$), (...}}\\
\end{tabular}
}
\end{center}

Without returning the entire webpage content, the browsing tool enables selective reading and iterative evidence extraction, which reduces context length pressure on the agent while maintaining information relevance.
\begin{figure*}[!t]
    \centering
    \includegraphics[width=2\columnwidth]{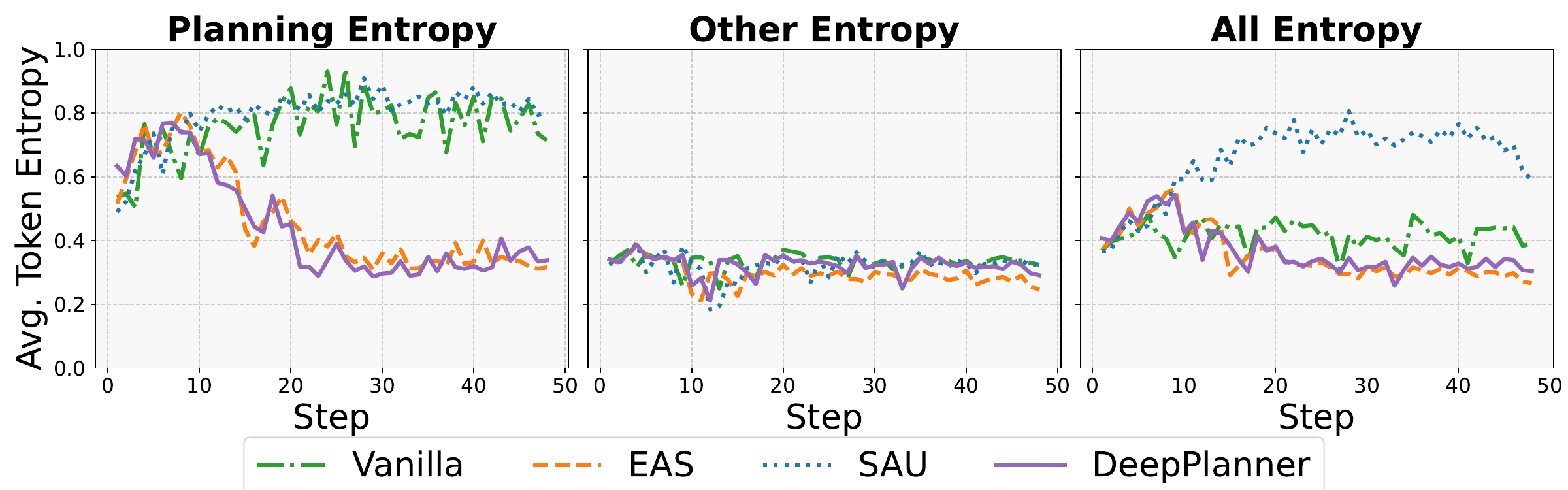}
    \caption{In the vanilla GRPO framework, planning-stage entropy is markedly higher than other execution-stage entropy, revealing underdeveloped planning capacity~\cite{cui2025entropymechanismreinforcementlearning}. With the advantage shaping mechanisms, our \papertitle effectively transforms the planning entropy to the performance on deep research tasks without inducing global entropy collapse.}
    \label{figs:entropy_trends}
    \vspace{-0.1in}
\end{figure*}

\section{\papertitle: Advantage Shaping}
In this section, we provide a detailed introduction to our end-to-end training framework based on GRPO~\cite{shao2024deepseekmathpushinglimitsmathematical}, with our proposed advantage shaping methodology, which enables more efficient planning optimization and high-quality rollout utilization.

\subsection{Vanilla RL Training Framework}
We train the policy $\pi_\theta$ of \papertitle using reinforcement learning under the GRPO framework.

\paragraph{GRPO} 
Unlike traditional sample-level loss functions~\cite{deepseekai2025deepseekr1incentivizingreasoningcapability}, this paper's GRPO operates at the token level~\cite{yu2025dapoopensourcellmreinforcement}, which better preserves individual token contributions in long text generation, where sample-level methods often dilute their impact. Formally, the objective function is given by:
\begin{equation}
\begin{aligned}
\small
\mathcal{J}(\theta) = &\mathbb{E}_{x \sim \mathcal{D}, \{y_i\}_{i=1}^G \sim \pi_{\theta_{\text{old}}}(\cdot|x)}  
    \frac{1}{\sum_{i=1}^G |y_i|}
    \sum_{i=1}^G \sum_{t=1}^{|y_i|}\\&
        \big[ \min \big( r_{i,t} A_{i,t}, \; \text{clip}(\cdot) A_{i,t} \big) 
        - \beta D_{\text{KL}} \big].
\end{aligned}
\end{equation}Here, $G$ denotes the number of rollouts $\tau$ generated by the agent for the same prompt. 
A rollout $y$ refers only to the tokens generated by the agent, while $\tau$ additionally includes tool responses. 
$|y_i|$ represents the length of the $i$-th rollout.
The importance sampling ratio is defined as
$
r_{i,t}(\theta) = 
\frac{\pi_{\theta}(a_{i,j,t} \mid s, a_{i,j,<t})}
     {\pi_{\text{old}}(a_{i,j,t} \mid s, a_{i,j,<t})}.
$
$D_{\text{KL}}$ measures the discrepancy between the current policy $\pi_\theta$ and the reference policy $\pi_{\text{old}}$. 
The clipping function is defined as $\text{clip}(1-\epsilon, r_{i,t}, 1+\epsilon)$, 
where $\epsilon$ and $\beta$ are hyperparameters. 
Finally, the group relative advantage is computed as
$
A_{i,j} = \frac{r_i - \mathrm{mean}(r)}{\mathrm{std}(r)},
$
which are computed using rollout rewards within the same group.

\paragraph{Rewards.}
The reward function is designed to balance correctness with adherence to the required output structure: (1) \textbf{Format Reward}: $+0.5$ if the output strictly follows the required structure (\texttt{<plan>}, \texttt{<tool\_call>}, \texttt{<answer>}). (2) \textbf{Answer Reward}: $+0.5$ if the final answer matches the ground truth. In this work, we employ an LLM-as-a-judge approach to determine the answer reward, as detailed in~\cref{app:evaluation_prompt}.
Importantly, format violations override correctness. 
For example, if the output format is incorrect (e.g., invalid tool call syntax or failing to generate a \texttt{<plan>} in the first round), the reward is set to 0, even if the final answer is correct.

\subsection{Entropy-based Advantage Shaping~(EAS)}
Under vanilla GRPO, as shown in~\cref{figs:entropy_trends}, planning tokens (\texttt{<think>} and \texttt{<plan>} segments in the planning steps) remain consistently higher entropy than other stages throughout training, indicating that the agent retains substantial uncertainty when forming plans.
Empirically, policy performance is fundamentally traded from policy entropy~\cite{cui2025entropymechanismreinforcementlearning}, making it critical to guide planning entropy toward a stable, reasonable range during training to improve downstream performance.
However, unlike RL tasks in mathematics~\cite{shao2024deepseekmathpushinglimitsmathematical} or coding, deep research training faces severe constraints due to long tool response times, making it impractical to scale training over hundreds of steps. Consequently, high-entropy planning receives insufficient advantage signals for optimization, leaving substantial room for exploration unexploited.
On the other hand, simply increasing the learning rate can accelerate convergence, but it also leads to \emph{entropy collapse}, where the model quickly converges to rigid patterns with minimal diversity and ceases to improve.
Inspired by ~\cite{cheng2025reasoningexplorationentropyperspective}, to accelerate the optimization of the planning capability and encourage exploration while preventing collapse, we add a gradient-detached entropy-based shaping term to the original token advantages:
\begin{equation}
\begin{aligned}
\small
\psi(\mathcal{H}_{i,t}) &= \min\left( 
    {\alpha} \cdot {\mathcal{H}_{i,t}^{\text{detach}}}, \; 
    \frac{|A_{i,t}|}{{\kappa}}
\right),\\
A^{\text{EAS}}_{i,t} &= A_{i,t} + \psi(\mathcal{H}_{i,t}),
\end{aligned}
\end{equation}
where $A_{i,t}$ is the original token advantage, 
$\mathcal{H}_{i,t}$ is the token-level entropy (detached from the computation
graph), $\alpha$ is a tunable shaping coefficient, and $\kappa \!>\! 1$ is a
clipping factor that prevents negative advantages from being flipped
positively, thereby preventing poor actions from being turned into favorable ones solely due to high entropy.

\subsection{Selective Advantage Upweighting~(SAU)}
Long-horizon tasks often contain multiple rollouts of varying difficulty.
Rollouts that are more complex tend to contribute more significantly to model improvement. Prior work \cite{zheng2025deepresearcherscalingdeepresearch} observes that the average number of tool calls increases gradually over the course of training. Building on this,~\cite{zhang2025evolvesearchiterativeselfevolvingsearch} proposed an iterative 
pipeline: filter rollouts that are both correct and belong to the top-$k$ in 
tool-call usage, apply supervised fine-tuning (SFT) on them, and then resume RL 
training. While effective, this approach introduces considerable engineering 
complexity. 

To achieve a similar and more efficient effect in an end-to-end manner, we introduce 
\emph{selective advantage upweighting}, which amplifies the advantages of 
high-quality, complex rollouts during RL training. This serves as an additional 
shaping strategy that mimics the benefits of SFT filtering while maintaining a 
simpler training pipeline. Furthermore, we optimize the filtering strategy used 
for rollout selection. As illustrated in \cref{figs:method}, within each 
rollout group, we select trajectories that: (1) achieve the maximum reward 
(1.0), and (2) use the minimum number of tool calls $N_{tool}$ in the group while 
exceeding a threshold ${c}$ controlling complexity. 
Formally, we define:
\begin{equation}
A^{\text{SAU}}_{i,t} = A_{i,t} \cdot \lambda, \quad N_{tool} \geq c,
\end{equation}
where $\lambda > 1$ controls the strength of upweighting. When both 
entropy-based and selective shaping are applied together in 
\papertitle, the update rule is:

\begin{equation}
\small
A^{\text{Shaping}}_{i,t} = 
\begin{cases} 
& A_{i,t} \cdot \lambda + \psi(\mathcal{H}_{i,t}), \text{if } \tau_i \text{ is selected} \\ 
&A_{i,t} + \psi(\mathcal{H}_{i,t}),  \text{if } \tau_i \text{ is not selected} 
\end{cases}
\end{equation}

Compared with previous methods that select rollouts purely based on tool-call counts, our strategy not only enforces a lower bound on rollout complexity but also promotes more efficient strategies, rather than simply encouraging excessive tool calls. This avoids over-rewarding redundant tool usage and reduces wasted resources.
\begin{table*}[t]
\renewcommand{\arraystretch}{1}
\small
\setlength{\tabcolsep}{6pt}
\begin{center}
\begin{tabular}{
    m{2.5cm}
    m{2.0cm}
    |m{0.4cm}<{\centering}m{0.4cm}<{\centering}m{0.7cm}<{\centering}m{0.5cm}<{\centering}m{0.5cm}<{\centering}
    |m{0.9cm}<{\centering}m{0.5cm}<{\centering}m{0.5cm}<{\centering}m{0.9cm}<{\centering}
    |m{0.9cm}<{\centering}
}
\toprule
\multirow{2}{*}{\textbf{Method}} &
\multirow{2}{*}{\textbf{\makecell[l]{Inference\\ Environment}}} &
\multicolumn{5}{c|}{In-domain} &
\multicolumn{4}{c|}{\textbf{Out-of-domain}} &
\multirow{2}{*}{\textbf{Total}} \\
& & NQ & TQ & Hotpot & 2Wiki & Avg & \textbf{Musique} & \textbf{Bamb} & \textbf{PopQA} & \textbf{Avg} & \\
\midrule
\multicolumn{11}{l}{\textbf{\textit{$\diamond$ Prompt Based}}} \\
CoT              & - & 32.0 & 48.2 & 27.9 & 27.3 & 33.9 & 7.4 & 21.6 & 15.0 & 14.7 & 25.7\\
CoT + RAG        & Local RAG & 59.6 & 75.8 & 43.8 & 24.8 & 51.0 & 10.0 & 27.2 & 48.8 & 28.7 & 41.4\\
Search-o1*       & Local RAG & 57.4 & 61.1 & 40.8 & 32.8 & 48.0 & 21.3 & 38.4 & 42.4 & 34.0 & 42.0\\
Search-o1        & Web Search & 55.1 & 69.5 & 42.4 & 37.7 & 51.2 & 19.7 & 53.6 & 43.4 & 38.9 & 45.9\\
\midrule
\multicolumn{11}{l}{$\clubsuit$ \textbf{\textit{Training Based}}} \\
Search-r1-base    & Local RAG & 60.0 & 76.2 & 63.0 & 47.9 & 61.8 & 27.5 & 57.6 & 47.0 & 44.0 & 54.2\\
Search-r1-instruct& Local RAG & 49.6 & 49.2 & 52.5 & 48.8 & 50.0 & 28.3 & 47.2 & 44.5 & 49.5 & 49.8\\
R1-Searcher       & Web Search & 52.3 & 79.1 & 53.1 & 65.8 & 62.6 & 25.6 & 65.6 & 43.4 & 44.9 & 55.0\\
DeepResearcher    & Web Search & 61.9 & 85.0 & 64.3 & 66.6 & 69.5 & 29.3 & 72.8 & 52.7 & 51.6 & 61.8\\
DeepResearcher*    & Web Search & 66.4 & 86.0 & 65.4 & 75.0& 73.2 & 29.0 & 71.7 & 50.2 & 50.3 & 63.4\\
EvolveSearch-ite1    & Web Search & 68.5 & 87.4 & 65.4 & 75.6 & 74.2 & 29.3 & 74.0 & 51.2 & 51.5 & 64.5\\
EvolveSearch-ite2    & Web Search & 69.4 & 86.3 & 66.3 & \textbf{78.5} & 75.1 & 31.6 & 76.5 & 52.8 & 53.6 & 65.9\\
EvolveSearch-ite3    & Web Search & 71.0 & \textbf{89.5} & \underline{67.7} & \underline{76.4} & \textbf{76.2} & \underline{33.8} & 77.1 & 50.3 & 53.7 & \underline{66.6}\\
\midrule
\multicolumn{11}{l}{$\spadesuit$ \textbf{\textit{Our Series}}} \\
\textbf{\papertitle}    & Web Search& 72.9 & \underline{87.5} & \textbf{68.2} & 75.6 & \underline{76.1} & 32.4 & \underline{77.6} & \textbf{55.3} & \underline{55.1} & \textbf{67.1}\\
- w/ \textit{EAS}    & Web Search& \underline{73.4}  & 85.9  & 66.6  & 72.1  & 74.5  & \textbf{34.4}  & \textbf{78.4}  & 54.5  & \textbf{55.8} & 66.5\\
- w/ \textit{SAU}    & Web Search& 70.3 & 86.7 & 67.0 & 64.8 & 72.2 & 33.4 & 76.0 & \underline{54.7} & 54.7 & 64.7\\
- \textit{Vanilla}    & Web Search& \textbf{73.8} & 87.3 & 65.8 & 68.8 & 73.9 & 30.1 & 71.2 & 53.7 & 51.7 & 64.4\\
\bottomrule
\end{tabular}
\end{center}
\caption{The overall performance across seven benchmarks shows that \papertitle achieves SOTA results in terms of the total average MBE score and significantly enhances the out-of-domain generalization. We also report DeepResearcher with model-based reward~\cite{zhang2025evolvesearchiterativeselfevolvingsearch} (denoted DeepResearcher*)}
\label{tabs:evaluation_overall_performance}
\end{table*}

\section{Experiments}
\subsection{Experiment Settings}
\paragraph{Dataset and Metrics}

We follow the training and evaluation protocol of DeepResearcher. The training set comprises NQ, TQ, HotpotQA, and 2Wiki in a 1:1:3:3 ratio, totaling 80,000 samples, with 75\% being multi-hop, matching the long-horizon information-seeking scenarios of deep research. For evaluation, as detailed in~\cref{app:evaluation_prompt}, we adopt model-based evaluation (MBE)~\cite{zhang2025evolvesearchiterativeselfevolvingsearch} using \texttt{chatgpt-4o-latest} as the judge to score final answers (correct = 1, incorrect = 0), and we report the average MBE on (1) In-domain: NQ, TQ, HotpotQA, and 2Wiki (2,048 examples). (2) Out-of-domain: Musique, Bamboogle, and PopQA (1{,}129 examples). Note that our RL training uses only a small subset~(3,072) of the entire training set, and~\cref {tabs:training_dataset} summarizes the training budgets for our method and the baselines.

\begin{figure}[!t]
    \centering
    \includegraphics[width=0.9\columnwidth]{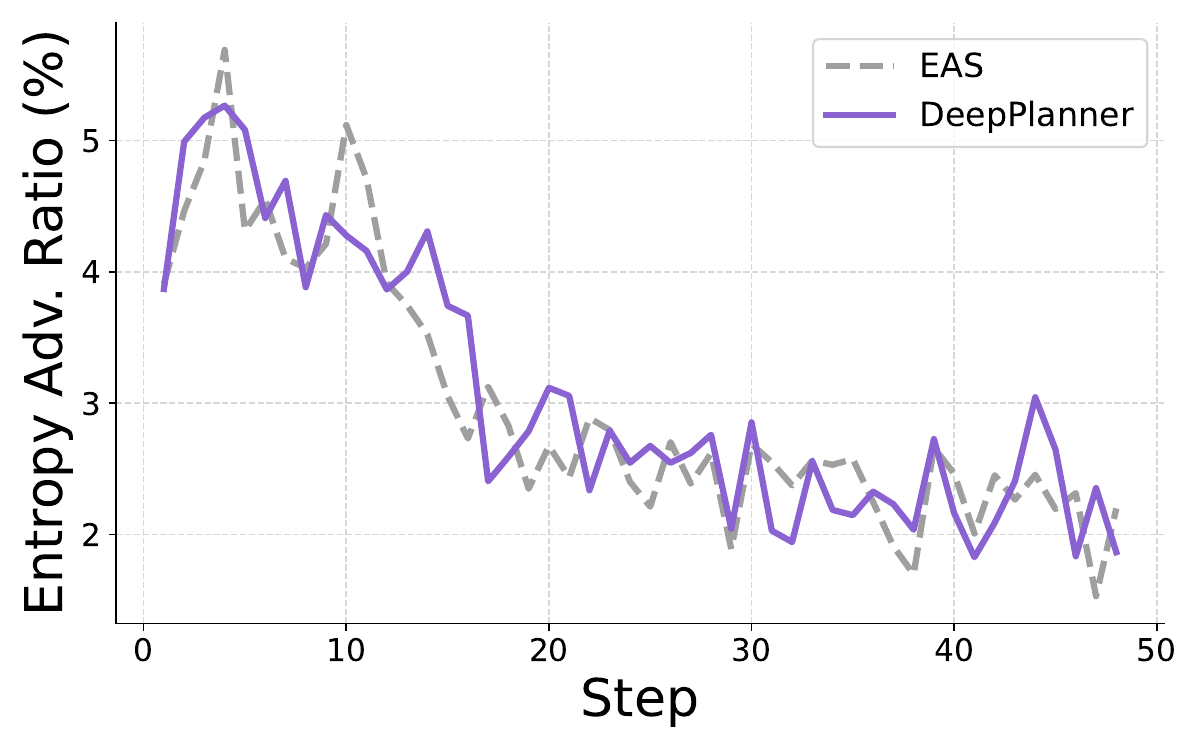}
    \caption{The ratio \( \frac{\mathcal{H}_{i,t}}{|A_{i,t}|} \) indicates that our approach does not over-encourage during the training process.}
    \label{figs:adv_ratio}
\end{figure}

\paragraph{Training Configuration}
In this paper, we choose \texttt{Qwen2.5-7B-Instruct} as the backbone model and the judge model during RL training. Each optimization step samples 64 queries and generates 8 rollouts for each query. For \papertitle and ablation studies, we all train for 48 steps.
For EAS, we set the shaping coefficient $\alpha = 0.1$ and the clipping factor $\kappa = 2$, following previous work~\cite{cheng2025reasoningexplorationentropyperspective}.
For SAU, we set $\lambda = 2$ and complexity threshold ${c} = 2$, as approximately 40\% of rollouts involve at least two tool calls.
Furthermore, we provide more implementation details in~\cref{app:implementation_details}.
In the final evaluation, we report four configurations for ablation: (1) Vanilla: GRPO without advantage shaping. (2) w/ EAS: GRPO with entropy-based shaping only. (3) w/ SAU: GRPO with selective upweighting only. (4) \papertitle: GRPO with both entropy-based shaping and selective upweighting.

\begin{table}[!t]
\centering
\renewcommand{\arraystretch}{1.2}
\setlength{\tabcolsep}{10pt}
\small
\begin{tabular}{lcc}
\toprule
\textbf{Method} & \textbf{\# Samples} & \textbf{\# Rollouts} \\
\midrule
DeepResearcher      & 8{,}000  & 16 \\
EvolveSearch-ite1   & 16{,}000 & 16 \\
EvolveSearch-ite2   & 24{,}000 & 16 \\
EvolveSearch-ite3   & 32{,}000 & 16 \\
\papertitle         & 3{,}072  & 8  \\
\bottomrule
\end{tabular}
\caption{Comparison of training sample size and rollout configurations across different methods.}
\label{tabs:training_dataset}
\vspace{-0.2in}
\end{table}

\paragraph{Baselines}
To evaluate the effectiveness of \papertitle, we compare against a series of baselines:
(1) CoT Only~\cite{wei2023chainofthoughtpromptingelicitsreasoning}: Chain-of-Thought prompting without external retrieval.
(2) RAG~\cite{gao2024retrievalaugmentedgenerationlargelanguage}: CoT with retrieved reference context for answer generation.
(3) Search-o1~\cite{li2025searcho1agenticsearchenhancedlarge}: A training-free baseline that sends search queries via APIs (e.g., Serper) and browses the webpage for richer evidence.
(5) Search-r1 (base/instruct)~\cite{jin2025searchr1trainingllmsreason}: RL-based framework trained with Wikipedia retrieval at train and inference time, using \textit{Qwen2.5-7B-base} and
\textit{Qwen2.5-7B-Instruct} as the backbone models.
(6) R1-Searcher~\cite{song2025r1searcherincentivizingsearchcapability}: Uses Bing search and answers by summarizing the first three result pages.
(7) DeepResearcher~\cite{zheng2025deepresearcherscalingdeepresearch}: Open-web deep research with autonomous URL selection rather than fixed top-3 webpage summarization.
(8) EvolveSearch~\cite{zhang2025evolvesearchiterativeselfevolvingsearch}: An RL–SFT loop using model-based reward. After each RL round, it filters the top-2,000 rollouts with the highest tool-call counts for SFT, then continues to the next RL round (four RL rounds and three SFT rounds in total).

\subsection{Overall Performance}
With fewer training samples and rollouts per sample, \papertitle achieves the best overall MBE: 67.1 with 3,072 samples and 8 rollouts, surpassing EvolveSearch-ite3 trained on 32,000 samples and 16 rollouts (see \cref{tabs:training_dataset,tabs:evaluation_overall_performance}). This underscores that scaling high-level planning quality, rather than merely scaling data or rollouts, is critical for improving deep research performance.
The entropy of planning token declines over training (Figure~\ref{figs:entropy_trends}), reflecting growing plan confidence. The ratio between the entropy-based shaping term and original advantage (Figure~\ref{figs:adv_ratio}) also drops, indicating our shaping avoids over-encouraging confident tokens, preventing premature entropy collapse while preserving exploration.

\subsection{Detailed Analysis}
\paragraph{Explicit planning matters.}
Analogous to how CoT elicits multi-step reasoning, explicitly requiring the agent to produce a global plan before acting yields structured, verifiable strategies. Compared with DeepResearcher$*$ without explicit planning, enforcing the \texttt{<plan>} stage improves MBE from 63.4 to 64.4, confirming that making intentions explicit stabilizes long-horizon behavior.

\paragraph{Format accuracy first boosts performance, and planning sustains gains.}
As illustrated in~\cref{figs:reward_trend_comparison}, the reward climbs rapidly around step 15, coinciding with a small entropy drop for tool-call and answer tokens, as well as a shift in the rollout distribution across reward tiers (reward < 0.5 indicates format error). The early performance surge is primarily due to better adherence to the required output structure. However, planning token entropy remains comparatively high at that point, and further reducing plan entropy correlates with sustained performance gains, which \papertitle and \textit{Entropy Adv} ablation achieve via entropy-based advantage shaping.

\begin{figure}[!t]
    \centering
    \includegraphics[width=1\columnwidth]{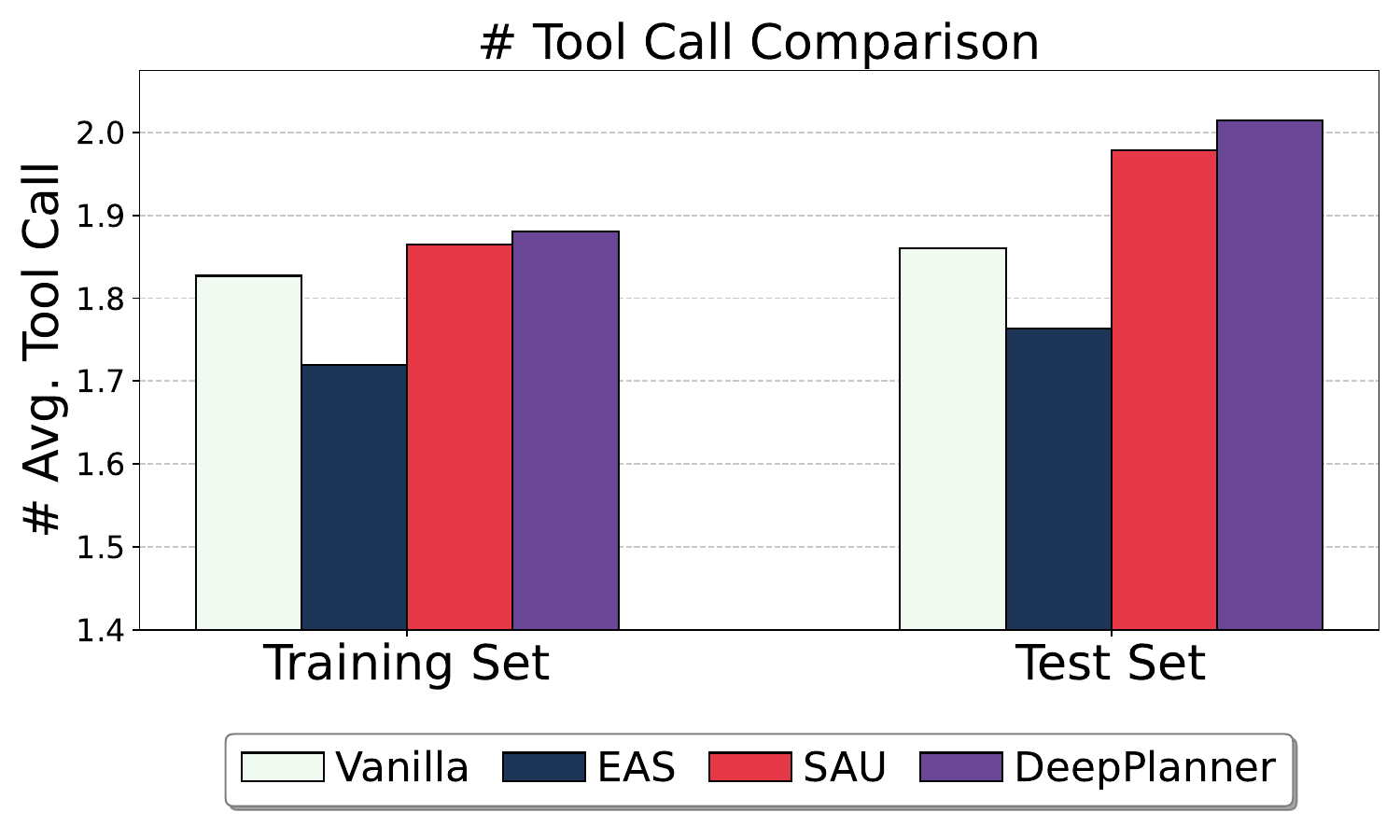}
    \caption{The average number of tool calls between different approaches in both the training and test sets.}
    \label{figs:tool_call_comparison}
    \vspace{-0.1in}
\end{figure}

\begin{figure}[!t]
    \centering
    \includegraphics[width=1\columnwidth]{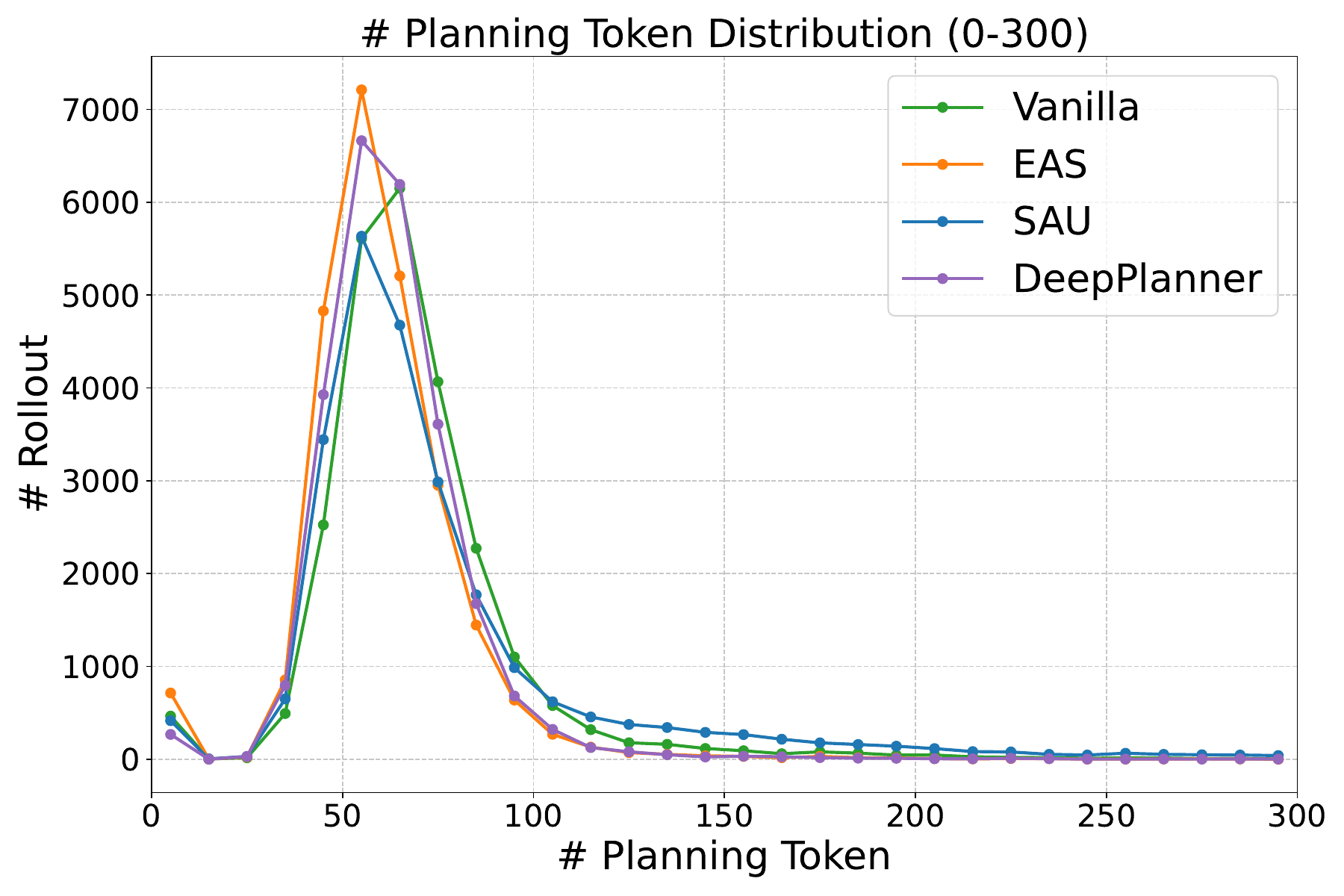}
    \caption{Distribution of planning tokens, with token length above 300 not reported.}
    \label{figs:planning_token_distribution}
    \vspace{-0.2in}
\end{figure}

\paragraph{Entropy-based shaping accelerates optimization while avoiding over-encouragement.}
Our entropy-based shaping amplifies learning signals on uncertain planning tokens, accelerating the discovery of effective strategies while clipping prevents flipping strongly negative advantages. Figures~\ref{figs:tool_call_comparison} and~\ref{figs:planning_token_distribution} indicate that the model learns concise, efficient plans under shaping. Unlike simply increasing the learning rate, our method preserves exploration and avoids entropy collapse (Figure~\ref{figs:entropy_trends}), preventing rigid, brittle behaviors.

\paragraph{Selective advantage upweighting encourages prudence by increasing tool calls.}
Selective advantage upweighting approximates the benefits of RL–SFT loops with minimal engineering effort.
Our selective mechanism encourages efficient, not excessive, tool use. Figure~\ref{figs:tool_call_comparison} shows a measured increase in tool calls where warranted, boosting in-domain performance from 74.5 to 76.1 and improving total performance. SAU alone does increase entropy (Figure~\ref{figs:entropy_trends}), reflecting increased uncertainty on harder problems and slower convergence. Nevertheless, it improves out-of-domain generalization (51.7 to 54.7), indicating robustness from targeted practice on complex cases.

\paragraph{EAS + SAU: faster learning on complex rollouts with stable plan convergence.}
Combining entropy-based shaping with selected rollouts focuses learning on the most impactful tokens within the most beneficial rollouts. The model simultaneously converges to reliable plans and learns to exercise prudence by allocating more tool calls on complex instances, yielding the best overall results (total 67.1).


\paragraph{Case Study}
As shown in~\cref{figs:case_study}, we remove the explicit planning stage from \papertitle, retrain under identical settings, and observe a typical short-sighted failure. With planning, the agent proposes a two-step plan (identify the father → find his birthplace), makes two focused \texttt{web\_search} calls, cross-checks sources, and returns the correct answer. Without planning, it only plans the first step, ignores global dependencies, mixes goals with missing intermediates, drifts via name concatenation (e.g., to Evan O’Neill Kane), wastes calls, and outputs the wrong answer, underscoring why explicit planning stabilizes entity resolution.

\section{Related Work}

\subsection{Deep Research Agents}
Early prompt-based search agents rely on fixed workflows to query and integrate external knowledge.
Systems such as OpenResearcher~\cite{zheng2024openresearcherunleashingaiaccelerated}, AirRAG~\cite{feng2025airragautonomousstrategicplanning}, IterDRAG~\cite{yue2025inferencescalinglongcontextretrieval}, Search-o1~\cite{li2025searcho1agenticsearchenhancedlarge}, and Open Deep Search~\cite{alzubi2025opendeepsearchdemocratizing} enhance search capabilities through carefully crafted prompts and interaction patterns, but their hand-engineered designs limit adaptability and generalization.
To overcome these limits, SFT-based approaches~\cite{yu2024autoragautonomousretrievalaugmentedgeneration} learn more flexible retrieval and synthesis policies. For example, CoRAG~\cite{wang2024coragcostconstrainedretrievaloptimization} couples SFT with MCTS to dynamically select document blocks under budgets, trading off compute at planning time and sensitivity to supervised signals.
RL-based methods utilize final outcome-supervised optimization to train end-to-end research agents, pushing the frontier of autonomous deep research capability, as demonstrated by ReSearch~\cite{chen2025researchlearningreasonsearch}, R1-Searcher~\cite{song2025r1searcherincentivizingsearchcapability}, Search-R1~\cite{jin2025searchr1trainingllmsreason}, WebRL~\cite{qi2025webrltrainingllmweb}, WebThinker~\cite{li2025webthinkerempoweringlargereasoning}, WebAgent-RL~\cite{wei2025webagentr1trainingwebagents}, DeepResearcher~\cite{zheng2025deepresearcherscalingdeepresearch}, and EvolveSearch~\cite{zhang2025evolvesearchiterativeselfevolvingsearch}.
Across these paradigms, planning is central: for complex problems, task decomposition typically helps LLMs execute more accurately and transparently. Plan*RAG~\cite{verma2025planragefficienttesttimeplanning} employs a separate model to produce an explicit plan and verifies its gains for information retrieval. Cognitive Kernel-Pro~\cite{fang2025cognitivekernelproframeworkdeep} introduces a planner that maintains a structured to-do list and a completed list, and DeepResearcher~\cite{zheng2025deepresearcherscalingdeepresearch} observes the emergence of planning ability during RL. In this paper, we provide the first systematic analysis of how planning influences RL-based deep research agents and introduce an end-to-end method that scales their planning abilities via advantage shaping~\cite{cheng2025reasoningexplorationentropyperspective}.

\subsection{Planning Capability of LLMs}
The planning capability of LLMs~\cite{wei2025plangenllmsmodernsurveyllm} to decompose high-level goals into actionable, temporally coordinated steps emerges as a central component of agentic systems.
Broadly, existing approaches fall into two paradigms:
(1) prompting LLMs to produce plans directly, which are then executed or lightly post-processed by downstream systems~\cite{wei2023chainofthoughtpromptingelicitsreasoning,wang2023planandsolvepromptingimprovingzeroshot,qin2023toolllmfacilitatinglargelanguage,liang2023taskmatrixaicompletingtasksconnecting,ahn2022icanisay,guo2023learningplannaturallanguage,zheng2024naturalplanbenchmarkingllms};
and (2) using LLMs to draft intermediate plans that are subsequently verified, refined, or expanded by symbolic planners, specialized agents, or external tools~\cite{liu2023llmpempoweringlargelanguage,singh2022progpromptgeneratingsituatedrobot,yuan2023skillreinforcementlearningplanning,kambhampati2024llmscantplanhelp,li2025hiplanhierarchicalplanningllmbased}.
In research-intensive, open-ended settings, systems such as~\cite{openaisdk,fang2025cognitivekernelproframeworkdeep} exemplify the value of explicit planning by separating plan generation from action execution, thereby enabling interpretable, verifiable, and progress-tracking behaviors. To further explore the planning ability in deep research agents, we provide a systematic, token-level entropy analysis that reveals persistently high plan-stage entropy as a key bottleneck, and introduce an advantage-shaping method to concentrate learning on uncertain planning decisions and complex rollouts, thereby directly scaling the agent planning capacity.


\section{Conclusion}
In this paper, we identify persistently high entropy in planning tokens of deep research agents, revealing untapped optimization potential. To address this, we propose \papertitle with two advantage shaping mechanisms: entropy-guided token-level shaping that accelerates planning optimization while preventing collapse, and selective upweighting that prioritizes complex, high-quality rollouts. Our approach achieves state-of-the-art performance with lower training budgets, demonstrating that targeted advantage shaping effectively scales planning capability in deep research agents.
\section*{Limitations}
First, due to the high cost of end-to-end RL in deep research tasks, we cap \papertitle and all ablations at 48 RL steps for fair comparison, requiring 24 hours on 8×A100 (80GB) GPUs, around \$50 in LLM API spend (mostly webpage-browsing summaries) and \$50 in Serper API for web search. Planning-token entropy remains relatively high, indicating headroom to further scale exploration of planning capability.
Second, following EvolveSearch’s analysis of model-based judgments, we do not independently evaluate the judge’s reliability, and we directly adopt \texttt{chatgpt-4o-latest} as the evaluator for fair comparison with EvolveSearch~\cite{zhang2025evolvesearchiterativeselfevolvingsearch}.
Finally, our method focuses on advantage shaping without explicit plan-stage rewards. Alternative approaches using fine-grained, multi-dimensional process reward for planning (e.g., quality, feasibility, consistency, verifiability) represent a different research direction deserving exploration.

\section*{Ethics Statement}
To the best of our knowledge, the backbone model~\cite{qwen2.5} and datasets~\cite{zheng2025deepresearcherscalingdeepresearch} used in this work are open-source and legally permissible for research use. Our experiments employ LLMs\footnote{\url{https://openrouter.ai/}} and tools\footnote{\url{https://serper.dev/}} under their respective licenses, and we adhered to the terms of service for all APIs used in this work.

\section*{Acknowledgments}
The authors of this paper were supported by the ITSP Platform Research Project (ITS/189/23FP) from the Innovation and Technology Commission of Hong Kong SAR, China, and by the AoE (AoE/E-601/24-N), RIF (R6021-20), and GRF (16205322) from the Research Grants Council of Hong Kong SAR, China.
We also thank the Amazon Stores Foundational AI team for their support.
\bibliography{custom}
\appendix

\begin{figure*}[!t]
    \centering
    \includegraphics[width=1.9\columnwidth]{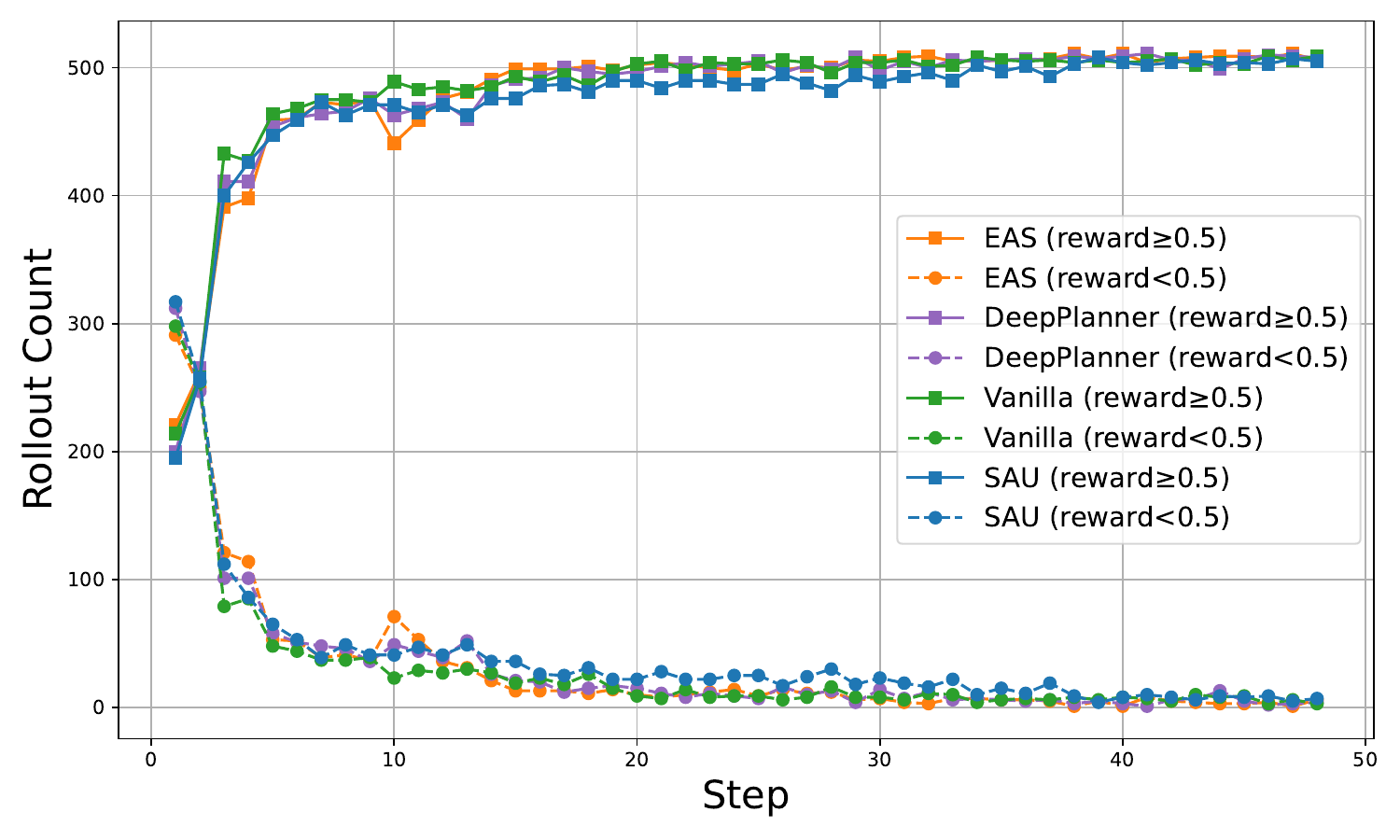}
    \caption{The average rollout reward over steps, where a reward less than 0.5 indicates a format error.}
    \label{figs:reward_trend_comparison}
\end{figure*}

\section{Implementation Details}
\label{app:implementation_details}

In this work, we implement \papertitle using Qwen2.5-7B-Instruct\footnote{\url{https://huggingface.co/Qwen/Qwen2.5-7B-Instruct}} as the backbone model. For agentic reinforcement learning, we adopt asynchronous rollouts using \texttt{agent\_loop} function\footnote{\url{https://verl.readthedocs.io/en/latest/advance/agent_loop.html}}of \texttt{VERL}~\cite{verl}.
Table \ref{tab:training_config} summarizes the hyperparameters used for training, and these values were maintained for all subsequent ablation experiments.

\begin{table}[h]
\centering
\caption{Training configuration.}
\label{tab:training_config}
\begin{tabular}{l l}
\hline
\textbf{Parameter} & \textbf{Value} \\
\hline
Training batch size (global) & 64 \\
Concurrent rollouts & 8 \\
Training steps & 48 \\
EAS coefficient \( \alpha \) & 0.1 \\
Clipping factor \( \kappa \) & 2 \\
SAU coefficient \( \beta \) & 2 \\
Complexity threshold \( c \) & 2 \\
KL loss coefficient & 0 \\
Entropy coefficient & 0 \\
Clip ratio \( \epsilon \) & 0.2 \\
Top-$k$ & $-1$ \\
Top-$p$ & $1$ \\
Temperature & $1$ \\
Maximum context length & 32{,}767 \\
\hline
\end{tabular}
\end{table}

For \papertitle and its three variants (Vanilla, EAS, and SAU only), we train via full-parameter fine-tuning on a single node with 8 NVIDIA A100 (80GB) GPUs; each run completes in approximately 24 hours. At evaluation time, we set the decoding temperature to 0 to perform deterministic greedy decoding.

\section{Details of \papertitle}

\subsection{Tool Schema}\label{app:tool_schema}
Figure \ref{tabs:tool_schema} presents the YAML Tool Schema Configuration employed by the \papertitle built upon the VERL architecture for external function calling. This configuration is essential for defining the interface and capabilities of the tools available to the agent.

\begin{figure*}[!t]
    \centering
    \includegraphics[width=2\columnwidth]{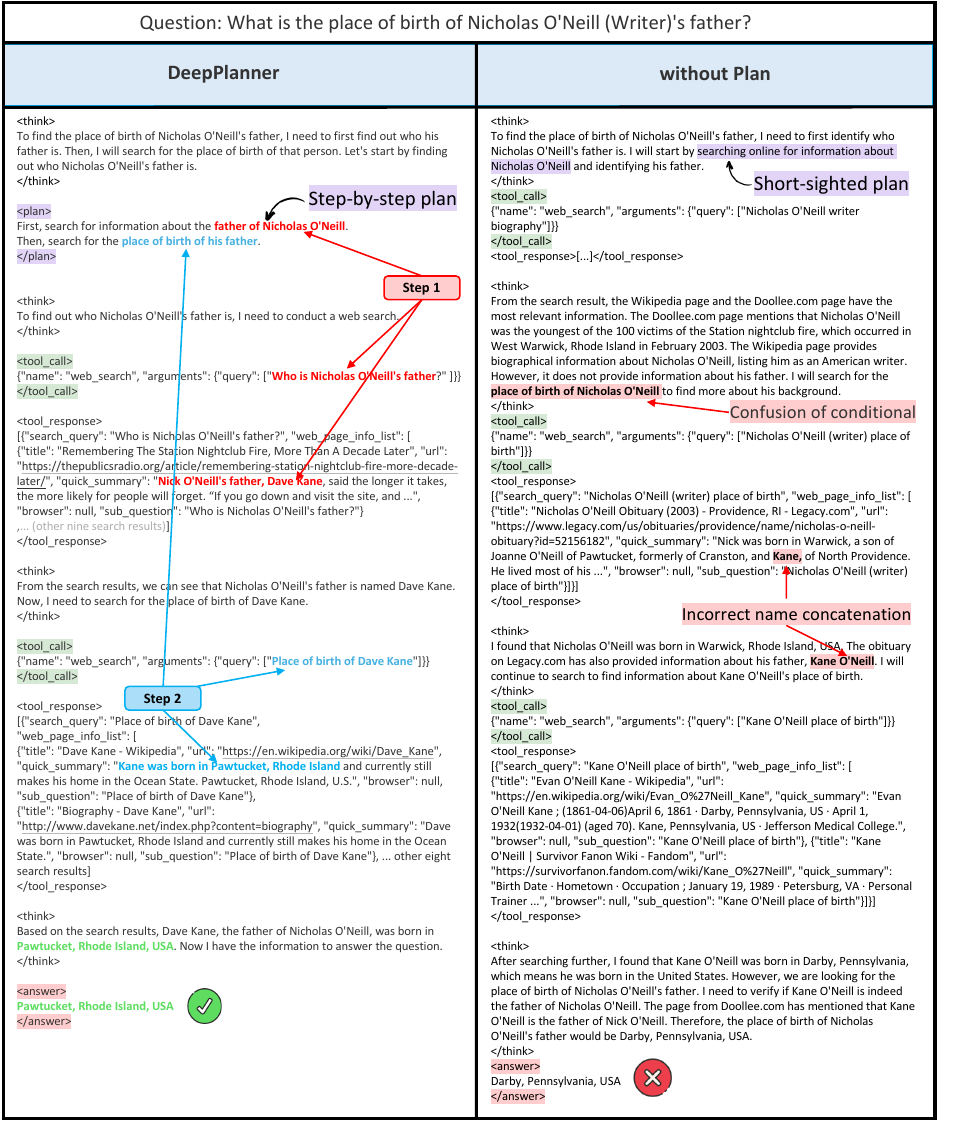}
    \caption{In the case study, we remove the explicit planning stage from \papertitle and retrain the model under the same conditions. The results reveal that without planning, the model fails to consider global dependencies, leading to errors such as confusing forgotten conditions and drifting to unrelated entities. This highlights that explicit planning is crucial for effective entity resolution and verification, as it prevents short-sighted behavior that results in retrieval biases and incorrect answers.}
    \label{figs:case_study}
\end{figure*}

\subsection{System Prompt of \papertitle}
\label{app:system_prompt}
\begin{figure*}[t]
\begin{tcolorbox}[
    colback = cBlue_1!5, 
    colframe = cBlue_6,  
    coltitle=white,
    fonttitle=\bfseries\small,
    fontupper=\small,
    fontlower=\small, 
    title=\textbf{\papertitle System Prompt}
]
\begin{lstlisting}[
    language=,
    breaklines=true,
    basicstyle=\ttfamily\small,
    columns=flexible,
    keepspaces=true,
    showstringspaces=false
]
## Background Information
* Today is {current_date}
* You are Deep Research AI Assistant, an expert in conducting thorough, multi-step research.

The question I give you is a complex question that requires a deep research to answer.

To help you perform this task, you are equipped with two tools:
- A web search tool to help you perform google search.
- A webpage browsing tool to help you get new page content.

## Your Task
Do not answer the question immediately.
In the first step, you must output your plan inside <plan></plan> tags.
In later steps, you can use <tool_call></tool_call> to call tools or <answer></answer> to provide your final answer.
You can also re-evaluate and update your plan during the later steps.

## Output Format
You must strictly follow one and only one of the three output formats below at each step:

<think>
Your thinking process here.
</think>
<plan>
Step-by-step research plan or re-plan. Each step should be concise and action-oriented.
</plan>

or

<think>
Your thinking process here.
</think>
<tool_call>
Tool call with correct format.
</tool_call>

or

<think>
Your thinking process here.
</think>
<answer>
Final answer only - a word, phrase, or number.
If it's a yes-or-no question, respond with only "yes" or "no"
No explanations or additional commentary.
</answer>
\end{lstlisting}
\end{tcolorbox}
\caption{The system prompt used to instruct the agent for complex multi-step research tasks. The agent is required to first provide its plan in the initial turn, and subsequently execute it strictly, including tool calls and the final answer. The agent is also permitted to modify its plan based on the evidence gathered in previous steps.}
\label{tabs:prompt_plan}
\end{figure*}

\begin{figure*}[t]
\begin{tcolorbox}[
    colback = cBlue_1!5, 
    colframe = cBlue_6,  
    coltitle=white,
    fonttitle=\bfseries\small,
    fontupper=\small,
    fontlower=\small, 
    title=\textbf{\papertitle System Prompt (Continue)}
]
\begin{lstlisting}[
    language=,
    breaklines=true,
    basicstyle=\ttfamily\small,
    columns=flexible,
    keepspaces=true,
    showstringspaces=false
]
# Tools

You may call one or more functions to assist with the user query.

You are provided with function signatures within <tools></tools> XML tags:
<tools>
{"type": "function", "function": {"name": "web_search", "description": "Search the web for relevant information from google. You should use this tool if the historical page content is not enough to answer the question. Or last search result is not relevant to the question.", "parameters": {"type": "object", "properties": {"query": {"type": "array", "description": "The queries to search"}}, "required": ["query"]}}}
{"type": "function", "function": {"name": "browse_webpage", "description": "Browse the webpage and return the content that not appeared in the conversation history. You should use this tool if the last action is search and the search result maybe relevant to the question.", "parameters": {"type": "object", "properties": {"url_list": {"type": "array", "description": "The chosen urls from the search result."}}, "required": ["url_list"]}}}
</tools>

For each function call, return a json object with function name and arguments within <tool_call></tool_call> XML tags:
<tool_call>
{"name": <function-name>, "arguments": <args-json-object>}
</tool_call>
\end{lstlisting}
\end{tcolorbox}
\caption{The agent automatically appends the tool definitions (encased in \texttt{<tools>...</tools>}) to the base instructions of the system prompt.}
\label{tabs:prompt_plan_tool}
\end{figure*}
As shown in~\cref{tabs:prompt_plan} and~\cref{tabs:prompt_plan_tool}, the prompt equips the agent with web search and browsing tools and mandates a structured workflow. Crucially, the agent must begin by outputting a structured \texttt{<plan></plan>}. Subsequent steps require strict adherence to defined output formats for tool calls (\texttt{<tool\_call></tool\_call>}) and final answers (\texttt{<answer></answer>}). In addition, the agent retains flexibility by being allowed to re-evaluate and update its plan based on new evidence gathered during the research process.

\subsection{Prompt of Evaluation}
\label{app:evaluation_prompt}
\begin{figure*}[t]
\begin{tcolorbox}[
    colback = cBlue_1!5, 
    colframe = cBlue_6,  
    coltitle=white, 
    fonttitle=\bfseries\small,
    fontupper=\small,
    fontlower=\small, 
    title=\textbf{Prompt for Final Answer Evaluation}
]
\begin{lstlisting}[
    language=,
    breaklines=true,
    basicstyle=\ttfamily\small,
    columns=flexible,
    keepspaces=true,
    showstringspaces=false
]
You will be given a question and its ground truth answer list where each item can be a ground truth answer. Provided a pred_answer, you need to judge if the pred_answer correctly answers the question based on the ground truth answer list.
You should first give your rationale for the judgement, and then give your judgement result (i.e., correct or incorrect).

Here is the criteria for the judgement:
1. The pred_answer doesn't need to be exactly the same as any of the ground truth answers, but should be semantically same for the question.
2. Each item in the ground truth answer list can be viewed as a ground truth answer for the question, and the pred_answer should be semantically same to at least one of them.

question: {question}
ground truth answers: {gt_answer}
pred_answer: {pred_answer}

The output should in the following json format:

The output should in the following json format:
'''json
{
"rationale": "your rationale for the judgement, as a text",
"judgement": "your judgement result, can only be 'correct' or 'incorrect'"
}
'''
Your output:
\end{lstlisting}
\end{tcolorbox}             
\caption{The specialized prompt used to evaluate the final answer correctness of the deep research agent. 
When the final judgment is \textit{correct}, the corresponding MBE score is assigned a value of 1; otherwise, the MBE score is 0.}
\label{tabs:prompt_evaluation} 
\end{figure*}
We employ a high-fidelity evaluation prompt to assess the final answers of deep research agents, focusing on semantic correctness against established ground truths. As detailed in Figure \ref{tabs:prompt_evaluation}, the prompt tasks a judge with comparing the agent's predicted answer to a list of acceptable ground truth answers. The core evaluation criterion is semantic equivalence, not direct lexical match.
The evaluator is mandated to provide a structured JSON output containing a \texttt{"rationale"} and a binary \texttt{"judgement"} (\texttt{"correct"} or \texttt{"incorrect"}). This strict output directly feeds into our performance metrics: a \texttt{"correct"} judgment assigns a MBE score of 1, while an \texttt{"incorrect"} judgment assigns an MBE score of 0.

\definecolor{TanYellowBackground}{HTML}{FFFFFA} 
\definecolor{TanYellowFrame}{HTML}{FFECB3}
\definecolor{KeywordGreen}{RGB}{0,128,0}
\definecolor{KeywordRed}{RGB}{180,0,0}

\lstdefinelanguage{yaml}{
    keywords={tools, class_name, config, type, tool_schema, function, name, description, parameters, properties, query, items, required, minItems, uniqueItems, url_list,  object, array, string},
    keywordstyle=\color{KeywordGreen}\bfseries,
    sensitive=false,
    comment=[l]{\#},
    commentstyle=\color{gray}\ttfamily,
    stringstyle=\color{KeywordRed}\ttfamily,
    morestring=[b]',
    morestring=[b]"
}

\begin{figure*}[t]
\begin{tcolorbox}[
colback = TanYellowBackground,
colframe = TanYellowFrame,
coltitle= black,
fonttitle=\bfseries\small,
fontupper=\small,
fontlower=\small,
title=\textbf{\papertitle Tool Schema Yaml}]
\begin{lstlisting}[
    language=yaml,
    breaklines=true,
    basicstyle=\ttfamily\small,
    columns=flexible,
    keepspaces=true,
    showstringspaces=false
]
tools:
  - class_name: "user.tools.websearch_tool.WebSearchTool"
    config:
      type: native
    tool_schema:
      type: "function"
      function:
        name: "web_search"
        description: "Search the web for relevant information from google. You should use this tool if the historical page content is not enough to answer the question. Or last search result is not relevant to the question."
        parameters:
          type: "object"
          properties:
            query:
              type: "array"
              items:
                type: "string"
                description: "The query to search, which helps answer the question"
              description: "The queries to search"
          required: ["query"]
          minItems: 1
          uniqueItems: true

  - class_name: "user.tools.browse_tool.BrowseWebpageTool"
    config:
      type: native
    tool_schema:
      type: "function"
      function:
        name: "browse_webpage"
        description: "Browse the webpage and return the content that not appeared in the conversation history. You should use this tool if the last action is search and the search result maybe relevant to the question."
        parameters:
          type: "object"
          properties:
            url_list:
              type: "array"
              items:
                type: "string"
                description: "The chosen url from the search result, do not use url that not appeared in the search result"
              description: "The chosen urls from the search result."
          required: ["url_list"]
\end{lstlisting}
\end{tcolorbox}
\caption{The YAML tool schema configuration used by \papertitle under the VERL architecture.}
\label{tabs:tool_schema}
\end{figure*}

\section{Case Study}
In detail, as shown in~\cref{figs:case_study}, we remove the explicit planning stage from DeepPlanner (no longer requiring first-round (<plan> … </plan>), keeping only (<think> … </think>) and tool calls), retrain under same training set, rollout number, GRPO hyperparameters, and tool configurations, and select a representative failure caused by short‑sighted planning. As shown in the left side of the figure, DeepPlanner first proposes a two‑step plan: Step 1: identify the writer Nicholas O’Neill’s father; Step 2: query the father’s birthplace, then executes two targeted \texttt{web\_search} calls guided by \texttt{sub\_question} prompts and cross‑checks Wikipedia and biography pages, yielding the correct answer: Pawtucket, Rhode Island, USA. In contrast, without an explicitly planning stage, the model only “plans” the first step and fails to consider global dependencies, mixing the end goal with missing intermediates: it starts from the writer’s bio, triggers a brittle conditional chain, commits name concatenation errors (e.g., merging “Kane” with “O’Neill”), drifts to the unrelated entity “Evan O’Neill Kane,” performs redundant and ineffective \texttt{web\_search} calls, and outputs the wrong answer, Darby, Pennsylvania, USA. This contrast shows that explicit planning constrains the search space, stabilizes entity resolution, and clarifies verification. Without planning, short‑sighted behavior accumulates linking and retrieval biases, precisely the failure mode our training addresses via entropy‑shaped advantage on uncertain planning tokens and selective upweighting of high‑quality, low‑tool‑call rollouts.


\end{document}